\documentclass[10pt,twocolumn,letterpaper]{article}

\usepackage{cvpr}              

\usepackage{graphicx}
\usepackage{amsmath}
\usepackage{amssymb}
\usepackage{booktabs}
\usepackage{color}
\usepackage{xcolor}

\usepackage{multicol}
\usepackage{multirow}
\usepackage[noend]{algpseudocode}
\usepackage{algorithmicx,algorithm}
\usepackage{bm}
\usepackage{parcolumns}
\usepackage[accsupp]{axessibility}  

\usepackage[pagebackref,breaklinks,colorlinks]{hyperref}

\usepackage[capitalize]{cleveref}
\crefname{section}{Sec.}{Secs.}
\Crefname{section}{Section}{Sections}
\Crefname{table}{Table}{Tables}
\crefname{table}{Tab.}{Tabs.}

\def\cvprPaperID{12825} 
\def\confName{CVPR}
\def\confYear{2023}

\begin{document}

\title{BBDM: Image-to-image Translation with Brownian Bridge Diffusion Models}

\author{Bo Li, Kaitao Xue, Bin Liu\\
School of Mathematics and Information Science, Nanchang Hangkong University, Nanchang, China\\
\and 
Yu-Kun Lai\\
School of Computer Sciences and Informatics, Cardiff University, Cardiff, UK\\
}

\maketitle

\begin{abstract}
   Image-to-image translation is an important and challenging problem in computer vision and image processing. Diffusion models~(DM) have shown great potentials for high-quality image synthesis, and have gained competitive performance on the task of image-to-image translation. However, most of the existing diffusion models treat image-to-image translation as conditional generation processes, and suffer heavily from the gap between distinct domains. In this paper, a novel image-to-image translation method based on the Brownian Bridge Diffusion Model~(BBDM) is proposed, which models image-to-image translation as a stochastic Brownian bridge process, and learns the translation between two domains directly through the bidirectional diffusion process rather than a conditional generation process. To the best of our knowledge, it is the first work that proposes Brownian Bridge diffusion process for image-to-image translation. Experimental results on various benchmarks demonstrate that the proposed BBDM model achieves competitive performance through both visual inspection and measurable metrics. 
\end{abstract}

\section{Introduction}

Image-to-image translation~\cite{Pixel2Pixel} refers to building a mapping between two distinct image domains. Numerous problems in computer vision and graphics can be formulated as image-to-image translation problems, such as style transfer~\cite{chen2016fast, gatys2016image, huang2017arbitrary, luan2017deep}, 
semantic image synthesis~\cite{liu2019learning, SPADE, OASIS, tang2020dual, tang2020local, Pixel2PixelHD} and sketch-to-photo synthesis~\cite{Pixel2Pixel, CycleGAN, CDiffE}.

\begin{figure}[ht]
\begin{center}
    \includegraphics[width=1.0\linewidth]{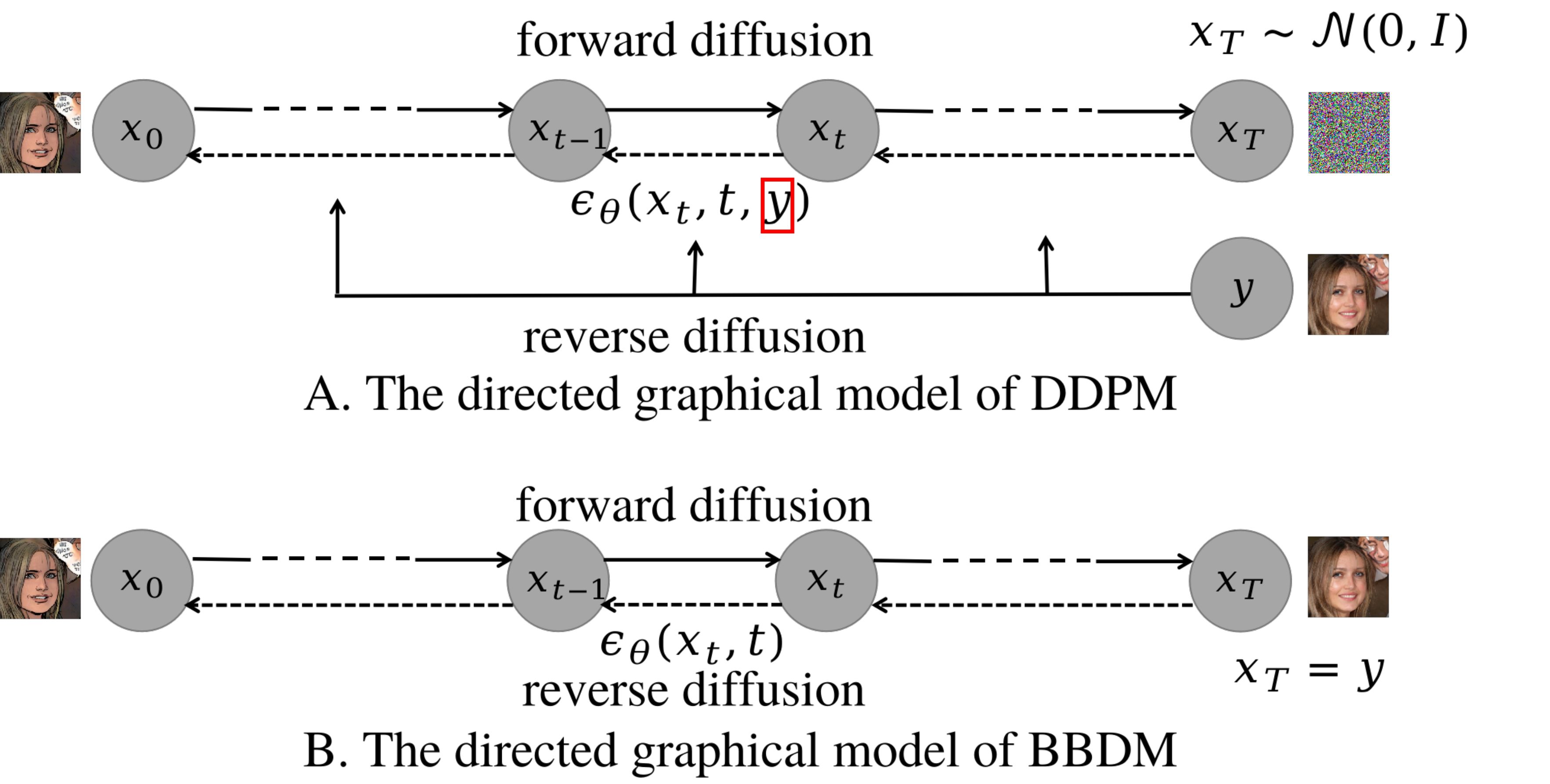}
\end{center}
   \caption{Comparison of directed graphical models of BBDM (Brownian Bridge Diffusion Model) and DDPM (Denoising Diffusion Probabilistic Model).}
\label{fig_BBDM_DDPM}
\end{figure}

A natural approach to image-to-image translation is to learn the conditional distribution of the target images given the samples from the input domain. Pix2Pix~\cite{Pixel2Pixel} is one of the most popular image-to-image translation methods. It is a typical conditional Generative Adversarial Network (GAN)~\cite{GAN}, and the domain translation is accomplished by learning a mapping from the input image to the output image. In addition, a specific adversarial loss function is also trained to constrain the domain mapping. Despite the high fidelity translation performance, they are notoriously hard to train~\cite{arjovsky2017wasserstein, gulrajani2017improved} and often drop modes in the output distribution~\cite{metz2016unrolled, ravuri2019classification}. In addition, most GAN-based image-to-image translation methods also suffer from the lack of diverse translation results since they typically model the task as a one-to-one mapping. Although other generative models such as Autoregressive Models~\cite{parmar2018image, van2016conditional}, VAEs (Variational Autoencoders)~\cite{kingma2013auto, vahdat2021deep}, and Normalizing Flows~\cite{dinh2016density, kingma2018glow} succeeded in some specific applications, they have not gained the same level of sample quality and general applicability as GANs.

\begin{figure*}[!ht]
\begin{center}
    \includegraphics[width=0.8\linewidth]{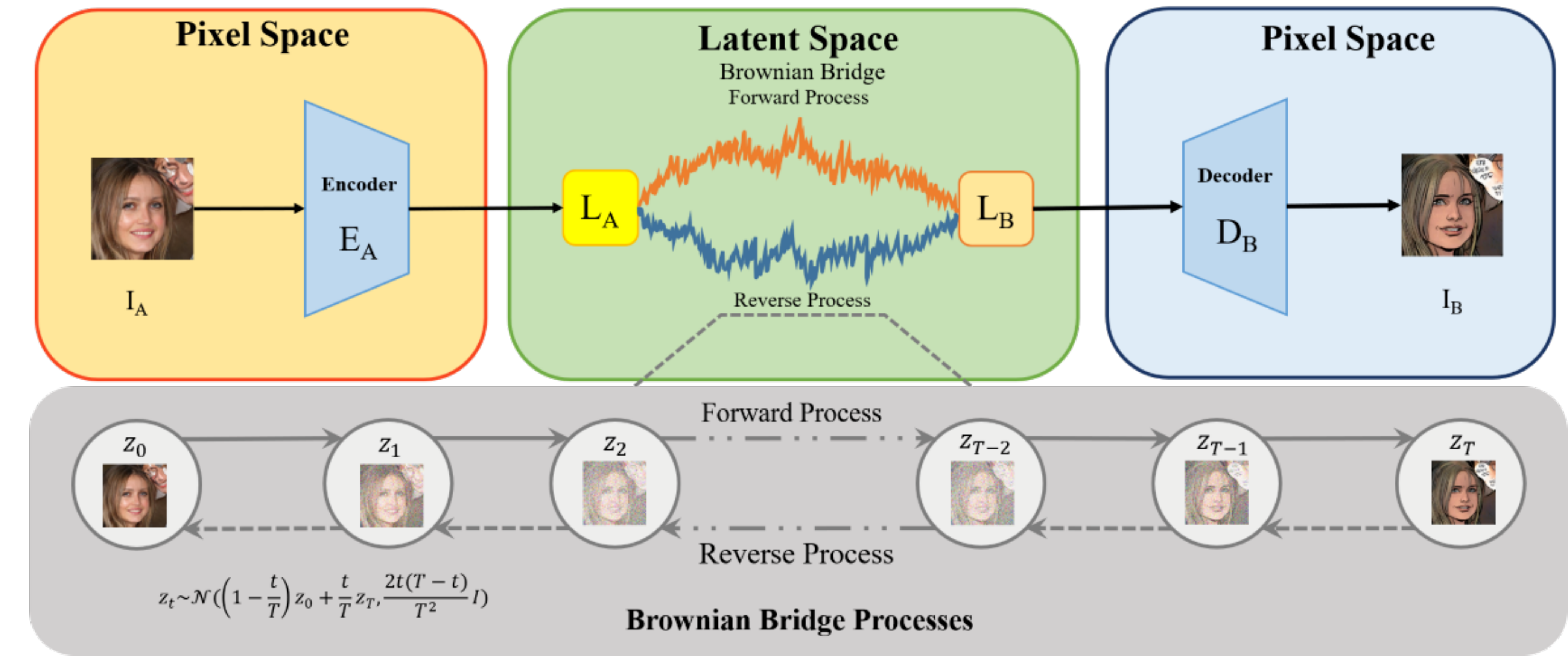}
\end{center}
   \caption{Architecture of BBDM.}
\label{fig_BBDM_architecture}
\end{figure*}

Recently, diffusion models~\cite{DDPM, sohl2015deep} have shown competitive performance on producing high-quality images compared with GAN-based models~\cite{GuidedDiffusion}. Several conditional diffusion models~\cite{CDiffE, ILVR, LDM, SR3, Palette} have been proposed for image-to-image translation tasks. These methods treat image-to-image translation as conditional image generation by integrating the encoded feature of the reference image into the U-Net in the reverse process~(the first row of Figure~\ref{fig_BBDM_DDPM}) to guide the diffusion towards the target domain. Despite some practical success, the above condition mechanism does not have a clear theoretical 
guarantee that the final diffusion result yields the desired conditional distribution.

Therefore, most of the conditional diffusion models suffer from poor model generalization, and can only be adapted to some specific applications where the conditional input has high similarity with the output, such as inpainting and super-resolution~\cite{CDiffE, ILVR, SR3}. Although LDM (Latent Diffusion Model)~\cite{LDM} improved the model generalization by conducting diffusion process in the latent space of certain pre-trained models, it is still a conditional generation process and the multi-modal condition is projected and entangled via a complex attention mechanism which makes LDM much more difficult to get such a theoretical guarantee. Meanwhile, the performance of LDM differs greatly across different levels of latent features showing instability. 

In this paper, we propose a novel image-to-image translation framework based on Brownian Bridge diffusion process. Compared with the existing diffusion methods, the proposed method directly builds the mapping between the input and the output domains through a Brownian Bridge stochastic process, rather than a conditional generation process. In order to speed up the training and inference process, we conduct the diffusion process in the same latent space as used in~LDM~\cite{LDM}. However, the proposed method differs from LDM inherently in the way the mapping between two image domains is modeled. The framework of BBDM is shown in the second row of Figure~\ref{fig_BBDM_DDPM}. It is easy to find that the reference image $\bm{y}$ sampled from domain $B$ is only set as the initial point $\bm{x}_T = \bm{y}$ of the reverse diffusion, and it will not be utilized as a conditional input in the prediction network $\bm{\mu}_{\theta}(\bm{x}_t, t)$ at each step as done in related works~\cite{CDiffE, ILVR, LDM, SR3}. The main contributions of this paper include:

\begin{enumerate}
    \item A novel image-to-image translation method based on Brownian Bridge diffusion process is proposed in this paper. As far as we know, it is the first work of Brownian Bridge diffusion process proposed for image-to-image translation. 
    
    \item The proposed method models image-to-image translation as a stochastic Brownian Bridge process, and learns the translation between two domains directly through the bidirectional diffusion process. The proposed method avoids the conditional information leverage existing in related work with conditional diffusion models.
    
    \item Quantitative and qualitative experiments demonstrate the proposed BBDM method achieves competitive performance on various image-to-image translation tasks.
\end{enumerate}

\section{Related Work}

In this section, we briefly review the related topics, including image-to-image translation,  diffusion models and Brownian Bridge.

\subsection{Image-to-image Translation}

 Isola \etal~\cite{Pixel2Pixel} firstly proposed a unified framework Pix2Pix for image-to-image translation based on conditional GANs. Wang \etal~\cite{Pixel2PixelHD} extended the Pix2Pix framework to generate high-resolution images. Unpaired translation methods like CycleGAN~\cite{CycleGAN} and DualGAN~\cite{DualGAN} used two GANs separately on two domains and trained them together with dual learning~\cite{DualLearning}, 
which allows them to learn from unpaired data.
However these one-to-one mapping translation methods fail to generate diverse outputs. With the aim of generating diverse samples, Lee \etal~\cite{DRIT} proposed DRIT++, but it requires that the condition image and result image must have high structural similarity. Several other GAN-based techniques have also been proposed for image-to-image translation such as unsupervised cross-domain method~\cite{taigman2016unsupervised}, multi-domain method~\cite{choi2018stargan}, few-shot method~\cite{liu2019few}. Nevertheless, GAN-based techniques suffer from the training instabilities and mode collapse problems. In addition to GAN-based models, Diffusion models~\cite{sohl2015deep} have also achieved impressive results on image generation~\cite{GuidedDiffusion, DDPM}, inpainting~\cite{Palette}, super-resolution~\cite{Palette, SR3}, text-to-image generation~\cite{LDM}.

\subsection{Diffusion Models}

 A $T$-step Denoising Diffusion Probabilistic Model(DDPM)~\cite{DDPM} consists of two processes: the forward process (also referred to as diffusion process), and the reverse inference process. 

The forward process from data $\bm{x}_0 \sim q_{data}(\bm{x}_0)$ to the latent variable $\bm{x}_T$ can be formulated as a fixed Markov chain:
\begin{align}
    q(\bm{x}_1, ..., \bm{x}_T | \bm{x}_0) = \prod_{t=1}^T q(\bm{x}_t | \bm{x}_{t-1}) \label{Eq-DDPMF}
\end{align}
where $q(\bm{x}_t | \bm{x}_{t-1}) = \mathcal{N}(\bm{x}_t; \sqrt{1-\beta_t}\bm{x}_{t-1}, \beta_t \bm{I})$ is a normal distribution, $\beta_t$ is a small positive constant. The forward process gradually perturbs $\bm{x}_0$ to a latent variable with an isotropic Gaussian distribution $p_{latent}(\bm{x}_T) = \mathcal{N}(\mathbf{0}, \bm{I})$. 

The reverse process strives to predict the original data $\bm{x}_0$ from the latent variable $\bm{x}_T \sim \mathcal{N}(\mathbf{0}, \bm{I})$ through another Markov chain:
\begin{align}
    p_{\theta}(\bm{x}_0, ..., \bm{x}_{T-1} | \bm{x}_T) =  \prod_{t=1}^T p_{\theta}(\bm{x}_{t-1} | \bm{x}_t) \label{Eq-DDPMR}
\end{align}
The training objective of DDPM is to optimize the Evidence Lower Bound~(ELBO). 
Finally, the objective can be simplified as to optimize:
\begin{align}
    \mathbb{E}_{\bm{x}_0, \bm{\epsilon}}||\bm{\epsilon}-\bm{\epsilon}_{\theta}(\bm{x}_t, t)||^2_2 \notag
\end{align}
where $\bm{\epsilon}$ is the Gaussian noise in $\bm{x}_t$ which is equivalent to $\triangledown_{\bm{x}_t}  \ln{q(\bm{x}_t | \bm{x}_0)} $, $\bm{\epsilon}_{\theta}$ is the model trained to estimate $\bm{\epsilon}$.

Most conditional diffusion models~\cite{CDiffE, ILVR, LDM, SR3, Palette} maintain the forward process and directly inject the condition into the training objective:
\begin{align}
    \mathbb{E}_{\bm{x}_0, \bm{\epsilon}}||\bm{\epsilon}-\bm{\epsilon}_{\theta}(\bm{x}_t, \bm{y}, t)||^2_2 \notag
\end{align}
Since $p(\bm{x}_t | y)$ dose not obviously appear in the training objective, it is difficult to guarantee the diffusion can finally reaches the desired conditional distribution.

Except for the conditioning mechanism, Latent Diffusion Model(LDM)~\cite{LDM} takes the diffusion and inference processes in the latent space of VQGAN~\cite{VQGAN}, which is proven to be more efficient and generalizable than operating on the original image pixels.

\subsection{Brownian Bridge}

A Brownian bridge is a continuous-time stochastic model in which the probability distribution during the diffusion process is conditioned on the starting and ending states.
Specifically, the state distribution at each time step of a Brownian bridge process starting from point $\bm{x}_0 \sim q_{data}(\bm{x}_0)$ at $t=0$ and ending at point $\bm{x}_T$ at $t=T$ can be formulated as:
\begin{align}
    p(\bm{x}_t | \bm{x}_0, \bm{x}_T) = \mathcal{N}\big((1-\frac{t}{T})\bm{x}_0 + \frac{t}{T}\bm{x}_T, \frac{t(T-t)}{T}\bm{I}\big) \label{Eq-OriginalBB}
\end{align}
It can be easily found that the process is tied down at both two ends with $\bm{x}_0$ and $\bm{x}_T$, and the process in between forms a bridge.

\section{Method} \label{section 3}

Given two datasets $\mathcal{X}_A$ and $\mathcal{X}_B$ sampled from domains $A$ and $B$, image-to-image translation aims to learn a mapping from domain $A$ to domain $B$. In this paper, a novel image-to-image translation method based on stochastic Brownian Bridge diffusion process is proposed. 
In order to improve the learning efficiency and model generalization, we propose to accomplish the diffusion process in the latent space of popular VQGAN~\cite{VQGAN}. The pipeline of the proposed method is shown in Figure \ref{fig_BBDM_architecture}.
Given an image $\mathbf{I}_{A}$ sampled from domain $A$, we can first extract the latent feature $\mathbf{L}_{A}$, and then the proposed Brownian Bridge process will map $\mathbf{L}_{A}$ to the corresponding latent representation $\mathbf{L}_{A \to B}$ in domain $B$. Finally, the translated image $\mathbf{I}_{A \to B}$ can be generated by the decoder of the pre-trained VQGAN. 

\subsection{Brownian Bridge Diffusion Model~(BBDM)} \label{section 3.1}

The forward diffusion process of DDPM~\cite{DDPM} starts from clean data $\bm{x}_0 \sim q_{data}(\bm{x}_0)$ and ends at a standard normal distribution. The setup of DDPM is suitable for image generation, as the reverse inference process naturally maps a sampled noise back to an image,
but it is not proper for the task of image translation between two different domains. Most of the existing diffusion-based image translation methods~\cite{CDiffE, ILVR, LDM, SR3} improved the original DDPM model by integrating the reference image as a conditional input in the reverse diffusion process.  

Different from the existing DDPM methods, a novel image-to-image translation method based on Brownian Bridge diffusion process is proposed in this section. 
Instead of ending at the pure Gaussian noise, Brownian Bridge process takes the clean conditional input $\bm{y}$ as its destination. We take similar notations as DDPM~\cite{DDPM}, and let $(\bm{x}, \bm{y})$ denote the paired training data from domains $A$ and $B$. To speed up the training and inference process, we conduct diffusion process in the latent space of popular VQGAN~\cite{VQGAN}. 
For simplicity and following notations as in DDPMs, we still use $\bm{x}, \bm{y}$ to denote the corresponding latent features $(\bm{x} := \mathbf{L}_A (\bm{x}), \bm{y} := \mathbf{L}_B (\bm{y}))$. The forward diffusion process of Brownian Bridge can be defined as:
\begin{gather}
    q_{BB}(\bm{x}_t | \bm{x}_0, \bm{y}) = \mathcal{N}(\bm{x}_t; (1-m_t)\bm{x}_0 + m_t\bm{y}, \delta_t\bm{I}) \label{Eq-BB} \\
    \bm{x}_0 = \bm{x}, \quad m_t = \frac{t}{T} \notag
\end{gather}
where $T$ is the total steps of the diffusion process, $\delta_t$ is the variance. It is noticed that if we take the variance of original Brownian Bridge as shown in Eq.(\ref{Eq-OriginalBB}),  $\delta_t = \frac{t(T-t)}{T}$, the maximum variance at the middle step, $\delta_{\frac{T}{2}} = \frac{T}{4}$, will be extremely large with the increase of $T$, and this phenomenon will  make the BBDM framework untrainable. Meanwhile, it has been mentioned in DDPM~\cite{DDPM} and VPSDE~\cite{song2020score} that the variance of middle steps should be preserved to be identity, if the distribution of $\bm{x}_0$ is supposed to be a standard normal distribution. Therefore, assuming that $\bm{x}_0, \bm{y} \sim \mathcal{N}(\mathbf{0}, \bm{I})$ are relatively independent, with the aim of preserving variances, a novel schedule of variance for Brownian Bridge diffusion process can be designed as
\begin{align}
     \delta_t &= 1 - \left((1 - m_t)^2 + m_t^2\right) \notag\\
              &= 2(m_t - m_t^2) \notag
\end{align}
It is easy to find that at the start of the diffusion process, i.e., $t=0$, we can have $m_0 = 0$, and the mean value is equal to $\bm{x}_0$ with probability 1 and variance $\delta_0=0$. When the diffusion process reaches the destination, $t=T$, we get $m_T = 1$, and the mean is equal to $\bm{y}$ while the variance $\delta_T = 0$. During the diffusion process, the variance $\delta_t$ will first grow  to the biggest value at the middle time $\delta_{max}= \delta_{\frac{T}{2}} = \frac{1}{2}$, and then it will drop until $\delta_T = 0$ at the destination of the diffusion.
According to the characteristic of Brownian Bridge diffusion process, the sampling diversity can be tuned by the maximum variance at the middle step $t = \frac{T}{2}$, therefore, we can scale $\delta_t$ by a factor $s$ to control the sampling diversity in practice: 
\begin{equation}
    \delta_t = 2s(m_t - m_t^2) \label{Eq-control_variance}
\end{equation}
We set $s=1$ 
by default,
and we will further discuss the influence of different $s$ values for sampling diversity in Section~\ref{section_ablation_study}.

\subsubsection{Forward Process}

According to the transition probability shown in Eq.(\ref{Eq-BB}), the forward diffusion of Brownian Bridge process only provides the marginal distribution at each step $t$. For training and inference purpose, we need to deduce the forward transition probability  $q_{BB}(\bm{x}_{t} | \bm{x}_{t-1}, \bm{y})$.

Given initial state $\bm{x}_0$ and destination state $\bm{y}$, the intermediate state $\bm{x}_t$
can be computed in discrete form as follows:
\begin{align}
    \bm{x}_t &= (1 - m_t)\bm{x}_0 + m_t \bm{y} + \sqrt{\delta_t} \bm{\epsilon}_t \label{Eq-xt}\\ 
     \bm{x}_{t-1} &= (1 - m_{t-1})\bm{x}_0 + m_t \bm{y} + \sqrt{\delta_{t-1}} \bf{\epsilon}_{t-1} \label{Eq-tminus}
\end{align}
 where $\bm{\epsilon}_t, \bm{\epsilon}_{t-1} \sim \mathcal{N}(\mathbf{0}, \bm{I})$. The transition probability $q_{BB}(\bm{x}_t | \bm{x}_{t-1}, \bm{y})$ can be derived by substituting the expression of $\bm{X}_0$ in Eq.(\ref{Eq-xt}) by the corresponding formula in Eq.(\ref{Eq-tminus})
\begin{align}
   q_{BB}(\bm{x}_t | \bm{x}_{t-1}, \bm{y}) = \mathcal{N}(\bm{x}_t; \frac{1-m_t}{1-m_{t-1}}\bm{x}_{t-1} \notag \\
    + (m_t - \frac{1-m_t}{1-m_{t-1}}m_{t-1})\bm{y}, \delta_{t|t-1}\bm{I}) \label{Eq-BBForwardMarkov}
\end{align}
where $\delta_{t|t-1}$ is calculated by $\delta_t$ as:
\begin{align}
    \delta_{t|t-1} = \delta_t - \delta_{t-1}\frac{(1-m_t)^2}{(1-m_{t-1})^2} \notag
\end{align}
According to Eq.(\ref{Eq-BBForwardMarkov}), when the diffusion process reaches the destination, i.e., $t = T$, we can get that $m_T = 1$ and $\bm{x}_T = \bm{y}$. The forward diffusion process defines a fixed mapping from domain $A$ to domain $B$.

\subsubsection{Reverse Process} \label{section3.1.2}

In the reverse process of traditional diffusion models, the diffusion process starts from a pure noise sampled from a Gaussian distribution, and eliminates the noise step by step to get the clean data distribution.  
In order to model the conditional distribution, the existing methods~\cite{CDiffE, ILVR, LDM, SR3} take the condition as an additional input of the neural network in the reverse diffusion process. 

Different from the existing diffusion-based image-to-image translation methods, the proposed Brownian Bridge process directly starts from the conditional input by setting $\bm{x}_T = \bm{y}$. Based on the main idea of denoising diffusion methods, the reverse process of the proposed method aims to predict $\bm{x}_{t-1}$ based on $\bm{x}_t$:
\begin{align}
    p_{\theta}(\bm{x}_{t-1} | \bm{x}_t, \bm{y}) = \mathcal{N}(\bm{x}_{t-1}; \bm{\mu}_{\theta}(\bm{x}_t, t), \Tilde{\delta}_t \bm{I})\label{reverse}
\end{align}
where $\bm{\mu}_{\theta}(\bm{x}_t, t)$ is the predicted mean value of the noise, and $\Tilde{\delta}_t$ is the variance of noise at each step.
Similar to DDPM~\cite{DDPM}, the mean value $\bm{\mu}_{\theta}(\bm{x}_t, t)$ is required to be learned by a neural network with parameters $\theta$ based on maximum likelihood criterion. 
Although the variance $\Tilde{\delta}_t$ does not need to be learned, it plays an important role in high-quality image translation.  
The analytical form of $\Tilde{\delta}_t$ will be introduced in Section~\ref{section3.1.3}. 

It is important to notice that the reference image $\bm{y}$ sampled from domain $B$ is only set as the start point $\bm{x}_T = \bm{y}$ of the reverse diffusion, and it will not be utilized as a conditional input in the prediction network $\bm{\mu}_{\theta}(\bm{x}_t, t)$ at each step as done in related works~\cite{CDiffE, ILVR, LDM, SR3}~(Figure~\ref{fig_BBDM_DDPM}).

\subsubsection{Training Objective} \label{section3.1.3}

The training process is performed by optimizing the Evidence Lower Bound~(ELBO) for the Brownian Bridge diffusion process which can be formulated as:
\begin{align}
    ELBO &= -\mathbb{E}_q\big(D_{KL}(q_{BB}(\bm{x}_T | \bm{x}_0, \bm{y}) || p(\bm{x}_T | \bm{y})) \notag \\
    &+ \sum^T_{t=2}D_{KL}(q_{BB}(\bm{x}_{t-1} | \bm{x}_t, \bm{x}_0, \bm{y}) || p_{\theta}(\bm{x}_{t-1} | \bm{x}_t, \bm{y})) \notag \\
    &- \log p_{\theta}(\bm{x}_0 | \bm{x}_1, \bm{y})\big)\label{ELBO}
\end{align}
Since $\bm{x}_T$ is equal to $\bm{y}$ in Brownian Bridge, the first term in Eq.(\ref{ELBO}) can be seen as a constant and ignored. By combining Eq.(\ref{Eq-BB}) and Eq.(\ref{Eq-BBForwardMarkov}), the formula $q_{BB}(\bm{x}_{t-1} | \bm{x}_t, \bm{x}_0, \bm{y})$ in the second term can be derived through Bayes' theorem and the Markov chain property:
\begin{align}
    q_{BB}(\bm{x}_{t-1} | \bm{x}_t, \bm{x}_0, \bm{y}) &= \frac{q_{BB}(\bm{x}_t | \bm{x}_{t-1}, \bm{y})q_{BB}(\bm{x}_{t-1} | \bm{x}_0, \bm{y})}{q_{BB}(\bm{x}_t | \bm{x}_0, \bm{y})} \notag \\
    & = \mathcal{N}(\bm{x}_{t-1}; \bm{\Tilde{\mu}}_t(\bm{x}_t, \bm{x}_0, \bm{y}) , \Tilde{\delta}_t \bm{I}) \label{Eq-BBReverseMarkov}
\end{align}
where the mean value term is:
\begin{align}
    \bm{\Tilde{\mu}}_t(\bm{x}_t, \bm{x}_0, \bm{y}) &= \frac{\delta_{t-1}}{\delta_t}\frac{1-m_t}{1-m_{t-1}}\bm{x}_t \notag \\
    & + (1-m_{t-1}\frac{\delta_{t|t-1}}{\delta_t})\bm{x}_0 \notag \\
    & + (m_{t-1} - m_t\frac{1-m_t}{1-m_{t-1}}\frac{\delta_{t-1}}{\delta_t})\bm{y} \label{Eq-BBMean}
\end{align}
and the variance term is:
\begin{align}
    \Tilde{\delta}_t = \frac{\delta_{t|t-1}\cdot\delta_{t-1}}{\delta_t} \label{Eq-BBVar}
\end{align}
As $\bm{x}_0$ is unknown in the inference stage, we propose to utilize a reparametrization method used in DDPM~\cite{DDPM} by combining Eq.(\ref{Eq-BB}) and Eq.(\ref{Eq-BBMean}). Then $\bm{\Tilde{\mu}}_t$ can be reformulated as:
\begin{align}
    \bm{\Tilde{\mu}}_t(\bm{x}_t, \bm{y}) = c_{xt} \bm{x}_t  + c_{yt} \bm{y} + c_{\epsilon t}\big(m_t(\bm{y}-\bm{x}_0) + \sqrt{\delta_t}\bm{\epsilon} \big) \notag
\end{align}
where 
\begin{align}
    c_{xt} &= \frac{\delta_{t-1}}{\delta_t}\frac{1-m_t}{1-m_{t-1}} + \frac{\delta_{t|t-1}}{\delta_t}(1-m_{t-1}) \notag \\
    c_{yt} &= m_{t-1} - m_t \frac{1-m_t}{1-m_{t-1}} \frac{\delta_{t-1}}{\delta_t} \notag \\
    c_{\epsilon t} &= (1 - m_{t-1}) \frac{\delta_{t|t-1}}{\delta_t} \notag
\end{align}
Instead of predicting the whole  $\bm{\Tilde{\mu}}_t$, we just train a neural network $\bf{\epsilon}_{\theta}$ to predict the noise.
For clarification, we can reformulate $\bm{\mu_{\theta}}$ in Eq.(\ref{reverse}) as a linear combination of $\bm{x}_t$, $\bm{y}$ and the estimated noise $\bm{\epsilon}_{\theta}$:
\begin{align}
    \bm{\mu_{\theta}}(\bm{x}_t, \bm{y}, t) = c_{xt}\bm{x}_t + c_{yt}\bm{y} + c_{\epsilon t}\bm{\epsilon}_{\theta}(\bm{x}_t, t) \label{Eqmu}
\end{align}
Therefore, the training objective ELBO in Eq.(\ref{ELBO}) can be simplified as:
\begin{align}
    \mathbb{E}_{\bm{x}_0, \bm{y}, \bm{\epsilon}}[c_{\epsilon t}||m_t(\bm{y}-\bm{x}_0) + \sqrt{\delta_t}\bm{\epsilon} - \bm{\epsilon}_{\theta}(\bm{x}_t, t)||^2] \notag
\end{align}

\subsection{Accelerated Sampling Processes}

Similar to the basic idea of DDIM \cite{DDIM}, the inference processes of BBDM can be accelerated by utilizing a non-Markovian process while keeping the same marginal distributions as Markovian inference processes. 

Now, given a sub-sequence of [1:$T$] of length $S$ $\lbrace \tau_1, \tau_2, ..., \tau_S \rbrace$, the inference process can be defined by a subset of the latent variables $\bm{x}_{1:T}$, which is $ \lbrace \bm{x}_{\tau_1}, \bm{x}_{\tau_2}, ..., \bm{x}_{\tau_S} \rbrace$, 
\begin{align}
    q_{BB}(\bm{x}_{\tau_{s-1}} | \bm{x}_{\tau{s}}, \bm{x}_0, \bm{y}) = \mathcal{N} \Big( (1 - m_{\tau_{s-1}})\bm{x}_0 + m_{\tau_{s-1}} \bm{y} + \notag \\
    \sqrt{\delta_{\tau_{s-1}} - \sigma_{\tau_s}^2}\frac{1}{\sqrt{\delta_{\tau_{s}}}}\big( \bm{x}_{\tau_s} - (1-m_{\tau_s})\bm{x}_0 - m_{\tau_s} \bm{y} \big), \sigma_{\tau_s}^2 \bm{I} \Big) \notag
\end{align}
A numerical experiment is conducted in Section~\ref{experi} to evaluate the performance with different numbers of sampling steps. To balance the sampling quality and efficiency, we choose $S = 200$ by default.  The whole training process and sampling process are summarized in Algorithm~\ref{al1} and Algorithm~\ref{al2}.

\begin{algorithm}[!t]
    \caption{Training}
    \begin{algorithmic}[1]
        \Repeat
        \State paired data $\bm{x}_0 \sim q(\bm{x}_0)$, $\bm{y} \sim q(\bm{y})$
        \State timestep $t \sim Uniform({1,...,T})$
        \State Gaussian noise $\mathbf{\epsilon} \sim \mathcal{N}(\mathbf{0}, \mathbf{I})$
        \State Forward diffusion
        $
            \bm{x}_t = (1 - m_t)\bm{x}_0 + m_t \bm{y} + \sqrt{\delta_t} \mathbf{\epsilon} \notag
        $
        \State Take gradient descent step on \par \qquad
        $            \mathbf{\triangledown}_{\theta}||m_t(\bm{y}-\bm{x}_0) + \sqrt{\delta_t}\mathbf{\epsilon} - \mathbf{\epsilon}_{\theta}\big(\bm{x}_t, t\big)||^2 \notag
        $
        \Until converged
    \end{algorithmic}\label{al1}
\end{algorithm}
\begin{algorithm}[!t]
    \caption{Sampling}
    \begin{algorithmic}[1]
        \State sample conditional input $\bm{x}_{T} = \bm{y} \sim q(\bm{y})$
        \For{$t$ = $T, \dots, 1$}
            \State $\bm{z} \sim \mathcal{N}(\mathbf{0}, \mathbf{I})$ if $t > 1$, else $\bm{z} = \mathbf{0}$
            \State $\bm{x}_{t-1} = c_{xt}\bm{x}_t + c_{yt}\bm{y} - c_{\epsilon t}\mathbf{\epsilon}_{\theta}(\bm{x}_t, t) + \sqrt{\Tilde{\delta}_t} \bm{z}$
        \EndFor
        \Return $\bm{x}_0$
    \end{algorithmic}\label{al2}
\end{algorithm}

\section{Experiments}\label{experi}

\begin{figure*}
\begin{center}
    \includegraphics[width=0.8\linewidth]{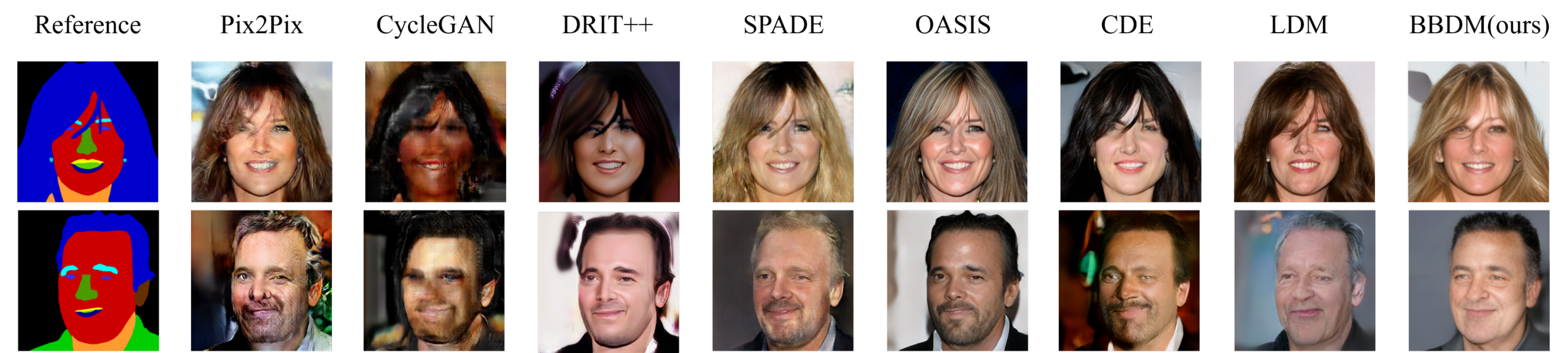} 
\end{center}
  \caption{Qualitative comparison on CelebAMask-HQ dataset.}
\label{fig_BBDM_quality_sample_CelebAMaskHQ}
\end{figure*}

\subsection{Experiment Setup} \label{section4.1}

\textbf{Models and hyperparameters}: The BBDM framework is composed of two components: pretrained VQGAN model and the proposed Brownian Bridge diffusion model. For fair comparison, We adopt the same pretrained VQGAN model as used in Latent Diffusion Model~\cite{LDM}. The number of time steps of Brownian Bridge is set to be 1000 during the training stage, and we use 200 sampling steps during the inference stage with the considerations of both sample quality and efficiency.

We train the network by using the Adam optimizer on a PC with an Intel Core i9-9900K CPU @ 3.2 GHz, 24GB RAM, and a GeForce GTX 3090 GPU.

\textbf{Evaluation}: For the visual quality and fidelity, we adopt the widely-used Fr\'echet Inception Distance (FID) and Learned Perceptual Image Patch Similarity (LPIPS) metrics~\cite{LPIPS}. To evaluate the generation diversity, we adopt the diversity metric proposed in \cite{CDiffE}. Specifically, we generate five samples $(\hat{\bm{x}}^5_{t=1})$ for a given conditional input $\bm{y}$, and calculate the average standard deviation for each pixel among the samples. Then, we report the average diversity over the whole test dataset.

\textbf{Datasets and baselines}: To demonstrate the capability of handling image-to-image translation on various datasets, We evaluate the BBDM framework on three distinct and challenging image-to-image translation tasks, including semantic synthesis task on CelebAMask-HQ dataset~\cite{CelebAMaskHQ}, sketch-to-photo task on edges2shoes and edges2handbags~\cite{Pixel2Pixel}, and style transfer task on faces2comics dataset. The baseline methods include Pix2Pix~\cite{Pixel2Pixel}, CycleGAN~\cite{CycleGAN}, DRIT++~\cite{DRIT}, CDE~\cite{SR3} and LDM~\cite{LDM}. Among the baselines, Pix2Pix, CycleGAN and DRIT++ are image-to-image translation methods based on conditional GANs, while CDE and LDM conduct image translation by conditional diffusion models. We additionally compare BBDM with OASIS~\cite{OASIS} and SPADE~\cite{SPADE} on CelebAMask-HQ dataset.

\begin{figure}[t]
\begin{center}
    \includegraphics[width=\linewidth]{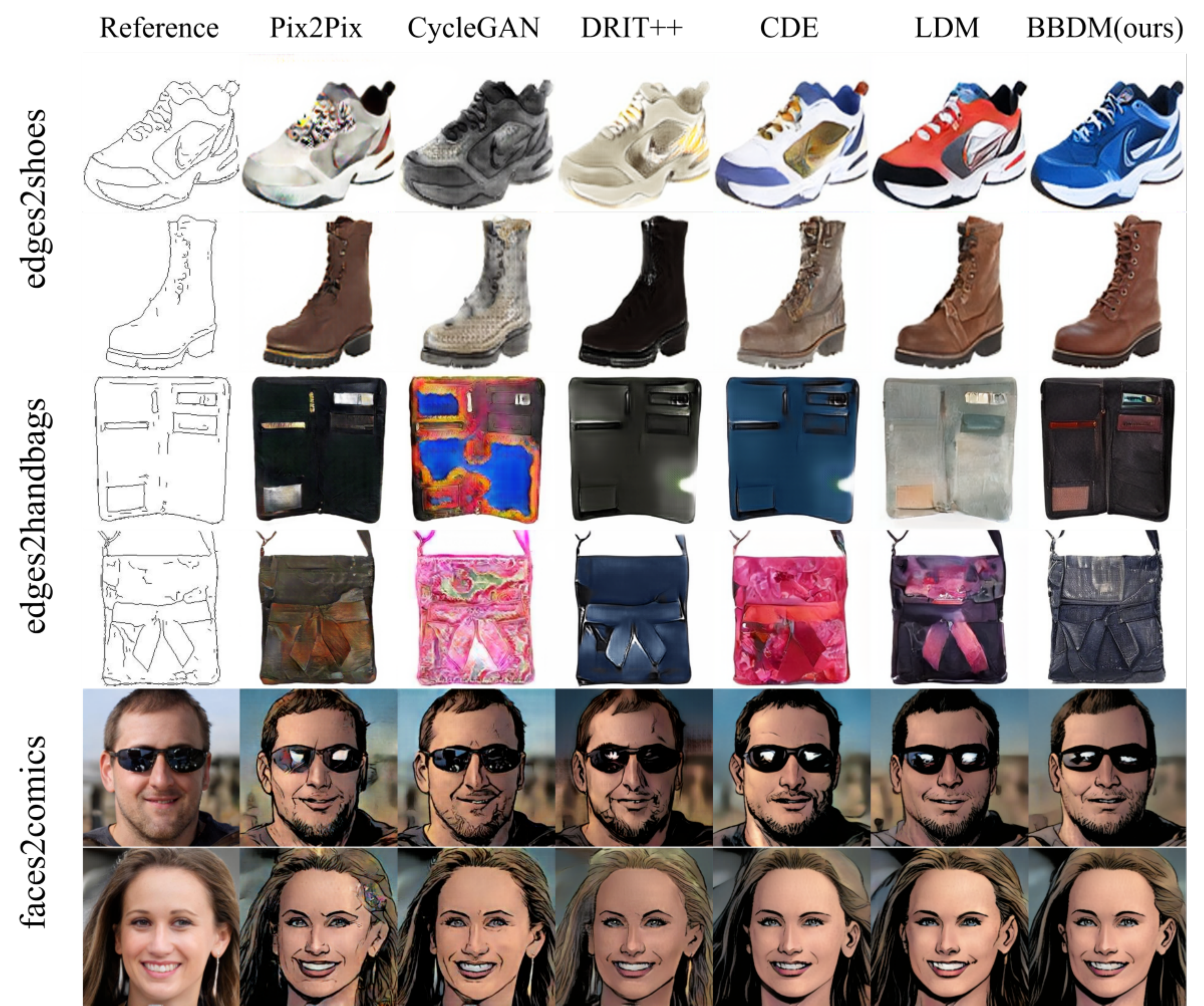} 
\end{center}
   \caption{Qualitative comparison on different image-to-image translation tasks.}
\label{fig_BBDM_quality_sample}
\end{figure}

\begin{figure}[t]
\begin{center}
    \includegraphics[width=\linewidth]{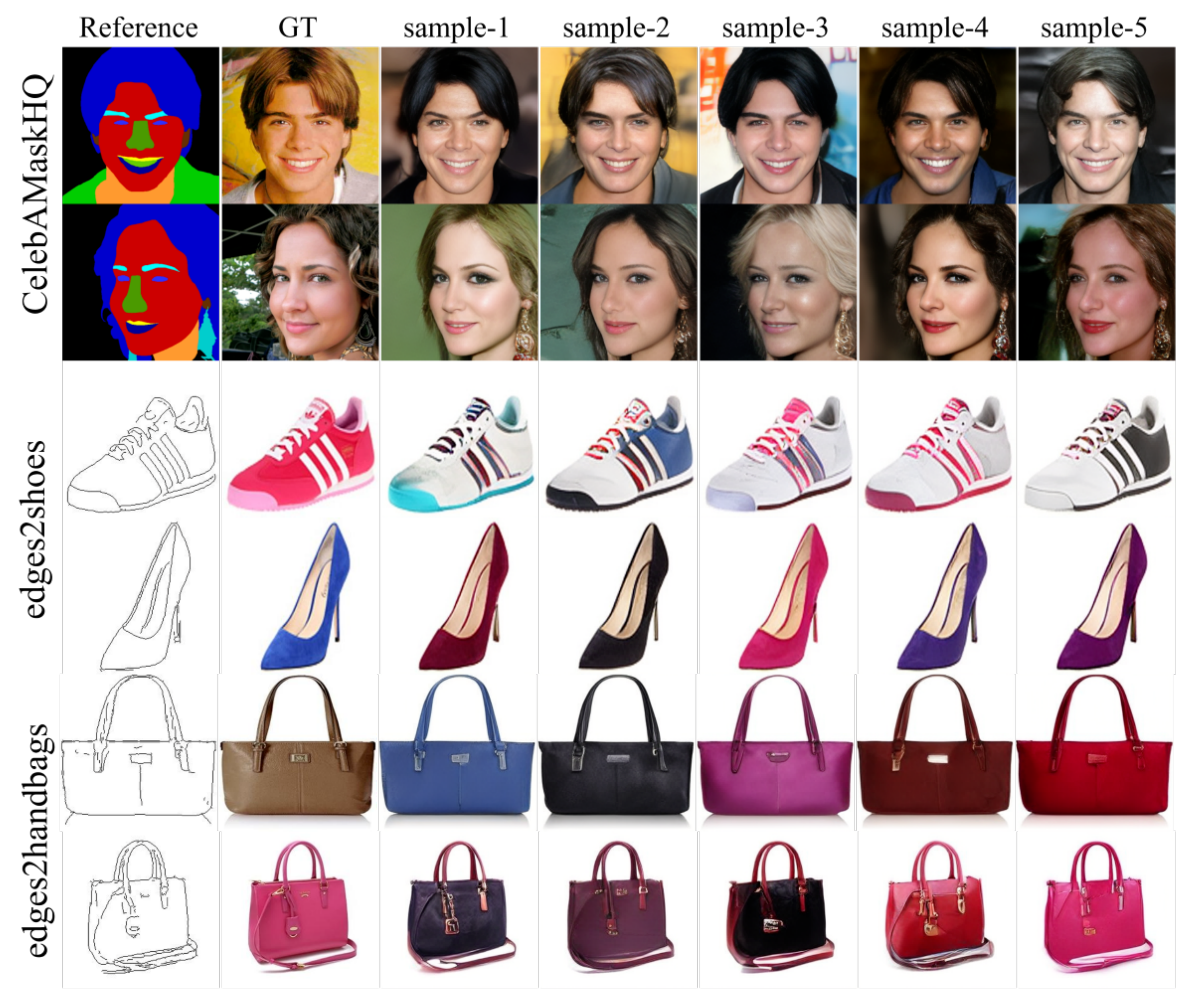}
\end{center}
   \caption{Diverse samples of BBDM on different image-to-image translation tasks.}
\label{fig_BBDM_diverse_sample}
\end{figure}

\subsection{Qualitative Comparison}

In this section, we evaluate the performance of the proposed BBDM against the state-of-the-art baselines on several popular image-to-image translation tasks. Semantic synthesis aims to generate photorealistic images based on semantic layout, while edge-to-image aims at synthesizing realistic image with the constraint of image edges. As both semantic layout and edge images are abstract, another task referred to as face-to-comics conducted on two domains with more similar distributions is involved.  

The experimental results of the proposed BBDM and other baselines are shown in  Figure~\ref{fig_BBDM_quality_sample_CelebAMaskHQ} and ~\ref{fig_BBDM_quality_sample}. Pix2Pix~\cite{Pixel2Pixel} can get reasonable results benefiting from the paired training data, while the performance of CycleGAN~\cite{CycleGAN} drops on small scale datasets. DRIT++ achieves better performance among GAN-based method, however, the translated images are oversmoothed and far from the ground truth distribution of the target domain. Compared with methods with GANs, diffusion based methods gain competitive performance. However, as is discussed in the introduction section, both CDE~\cite{SR3} and LDM~\cite{LDM} are conditional diffusion models, and suffer from conditional information leverage during the diffusion process. For example, when there are irregular occlusions as shown in the first row of Figure~\ref{fig_BBDM_quality_sample_CelebAMaskHQ}, CDE and LDM cannot generate satisfactory results due to the mechanism of integrating conditional input into the diffusion model. In contrast, the proposed BBDM conducts image-to-image translation by directly learning a diffusion process between these two domains, and avoids the conditional information leverage.

Benefiting from the stochastic property of Brownian Bridge, the proposed method can generate samples with high fidelity and diversity. Some examples are shown in Figure~\ref{fig_BBDM_diverse_sample}. 

\begin{table}[t]
    \centering
    \begin{tabular}{ccccc} \hline
        \multicolumn{1}{c}{\multirow{2}{*}{model}} & & \multicolumn{3}{c}{CelebAMask-HQ}\\ \cline{3-5}
        & & FID $\downarrow$ & LPIPS $\downarrow$ & Diversity $\uparrow$ \\ \cline{1-1} \cline{3-5}
        Pix2Pix & & 56.997 & 0.431 & 0 \\
        CycleGAN & & 78.234 & 0.490 & 0 \\
        DRIT++ & & 77.794 & 0.431 & 35.759 \\
        SPADE & & 44.171 & 0.376 & 0 \\
        OASIS & & 27.751 & 0.384 & 39.662 \\
        CDE & & 24.404 & 0.414 & \textbf{50.278} \\
        LDM & & 22.816 & 0.371 & 20.304 \\ \cline{1-1} \cline{3-5} 
        BBDM(ours) & & \textbf{21.350} & \textbf{0.370} & 29.859  \\ 
        \hline
    \end{tabular}
    \caption{Quantitative comparison on CelebAMask-HQ dataset.}
    \label{table_CelebAMaskHQ_quality_scores}
\end{table}

\begin{table*}[t]
    \setlength{\tabcolsep}{1.7mm}{
    \centering
    \begin{tabular}{ccccccccccccc} \hline
        \multicolumn{1}{c}{\multirow{2}{*}{model}} & & \multicolumn{3}{c}{edges2shoes} & & \multicolumn{3}{c}{edges2handbags} & & \multicolumn{3}{c}{faces2comics}\\ \cline{3-5} \cline{7-9} \cline{11-13}
        & & FID $\downarrow$ & LPIPS $\downarrow$ & Diversity $\uparrow$ & & FID $\downarrow$ & LPIPS $\downarrow$ & Diversity $\uparrow$  & & FID $\downarrow$ & LPIPS $\downarrow$ & Diversity $\uparrow$ \\ \cline{1-1} \cline{3-5} \cline{7-9} \cline{11-13}
        Pixel2Pixel & & 36.339 & 0.183 & 0 & & 32.994 & 0.273 & 0  & & 49.964 & 0.282 & 0 \\
        CycleGAN & & 66.115 & 0.276 & 0 & & 40.175 & 0.367 & 0  & & 35.133 & 0.263 & 0  \\
        DRIT++ & & 53.373 & 0.498 & \textbf{23.552} & & 43.675 & 0.411 & \textbf{30.169} & & 28.875 & 0.285 & 18.047 \\
        CDE & & 21.189 & 0.196 & 14.980 & & 28.575 & 0.313 & 24.158 & & 33.983 & 0.259 & \textbf{19.532}\\
        LDM & & 13.020 & \textbf{0.173} & 10.999 & & 24.251 & 0.307 & 22.705 & & 24.280 & 0.205 & 9.032\\ \cline{1-1} \cline{3-5} \cline{7-9} \cline{11-13}
        BBDM(ours) & & \textbf{10.924} & 0.183 & 12.226 & & \textbf{17.257} & \textbf{0.286} & 15.656 & & \textbf{23.203} & \textbf{0.192} & 10.046\\
        \hline
    \end{tabular}
    }
    \caption{Quantitative comparison on different image-to-image translation tasks.}
    \label{table_quality_scores}
\end{table*}

\subsection{Quantitative Comparison}
In this section, we compare the proposed BBDM against baselines with several popular quantitative metrics, including  FID,  
LPIPS and diversity measurement~\cite{CDiffE}. The numerical results are shown in Table~\ref{table_CelebAMaskHQ_quality_scores} and ~\ref{table_quality_scores}. It is obvious that the proposed BBDM method achieves the best FID performance on all of the four tasks, and gains competitive LPIPS scores.

\subsection{Other Translation Tasks}
\begin{figure}[!ht]
\begin{center}
    \includegraphics[width=0.8\linewidth]{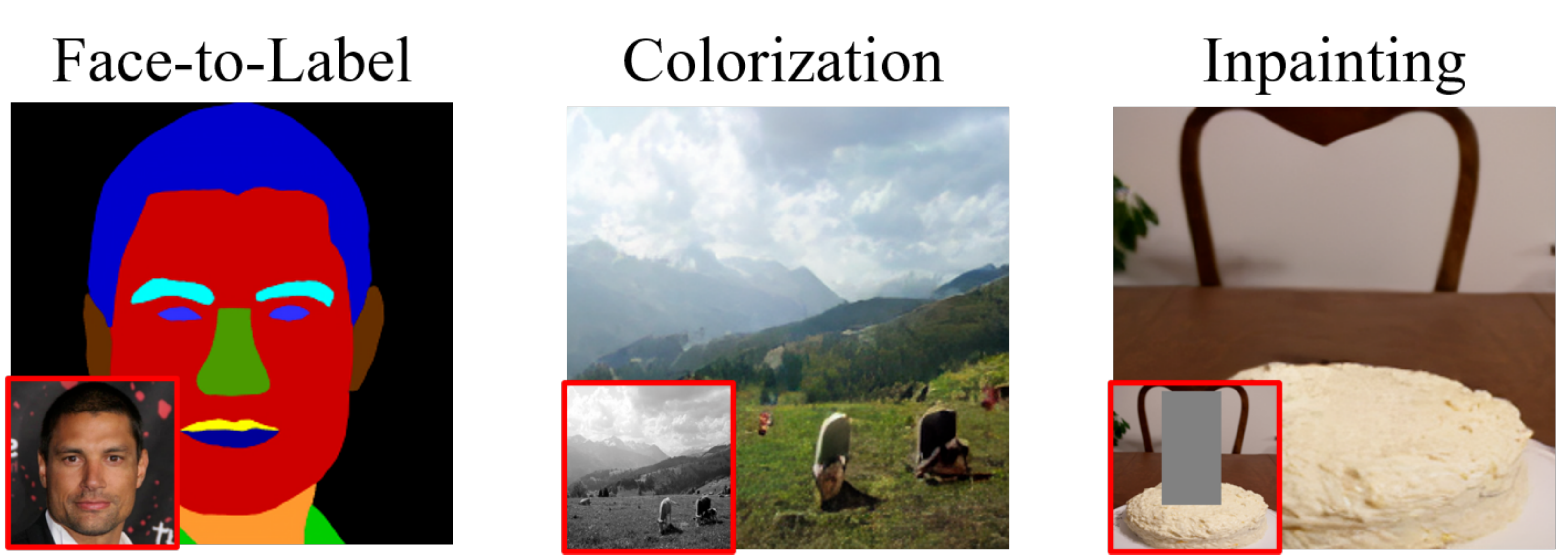}
\end{center}
   \caption{Face-to-label, colorization and inpainting results.}
\label{fig_other_task_6}
\end{figure}

In order to further verify the generalization of BBDM, we conducted inpainting, colorization experiments on VisualGENOME~\cite{visualGENOME} and face-to-label on CelebAMask-HQ~\cite{CelebAMaskHQ}. The experimental results in Figure~\ref{fig_other_task_6} show that BBDM can achieve comparable performance on various image translation tasks. More examples are shown in supplementary materials.

\subsection{Ablation Study} \label{section_ablation_study}

\begin{figure}[!t]
\begin{center}
    \includegraphics[width=1.0\linewidth]{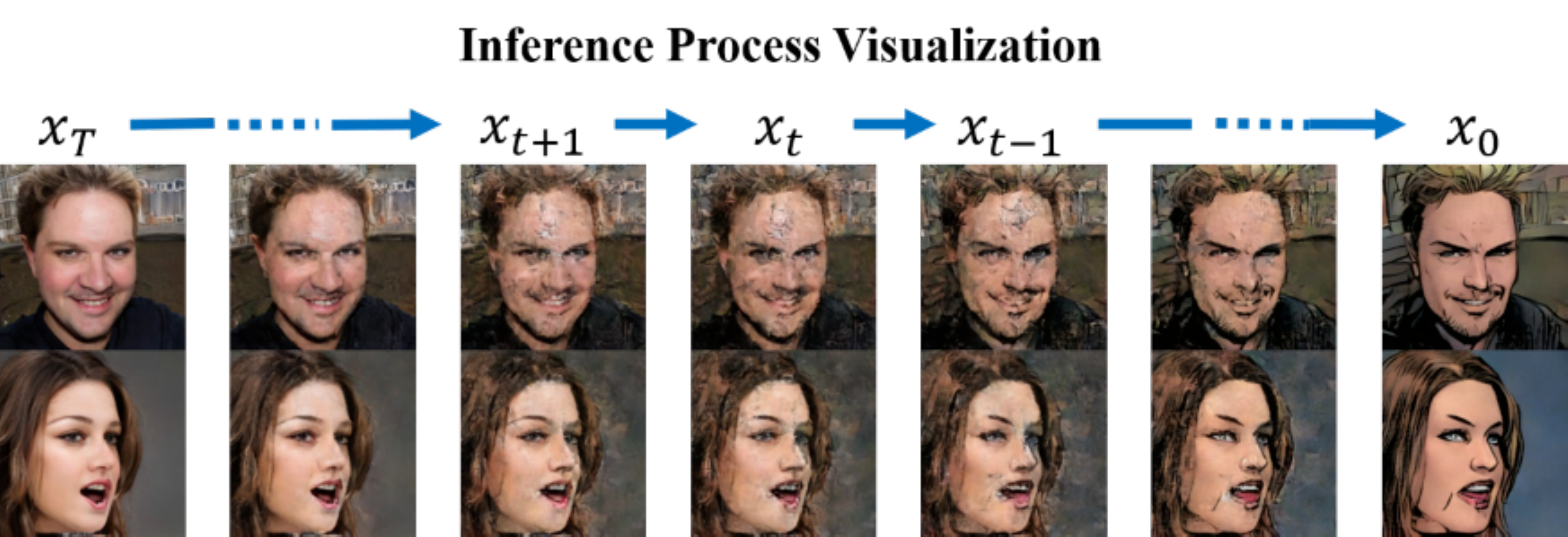}
\end{center}
   \caption{Latent space visualization.}
\label{fig_latent_visual}
\end{figure}
We perform ablative experiments to verify the effectiveness of several important designs in our framework.

\textbf{Influence of the pre-trained latent space}: To speed up the training and inference process, the diffusion process of the proposed BBDM is conducted in a pre-trained latent space the same as the one used in LDM~\cite{LDM}. In order to demonstrate the influence of different latent spaces to the performance of the proposed method, we conduct an ablation study by choosing different downsampling factors for VQGAN model as done in LDM.

In this experiment, we compare our BBDM framework and LDM with downsampling factors $f \in \{4, 8, 16\}$ on CelebAMask-HQ. For fair comparison, We implemented BBDM based on the same network structure as LDM and used the same VQGAN-f4, VQGAN-f8, VQGAN-f16 checkpoints of LDM. The quantitative metrics are shown in Table~\ref{table_LDM_BBDM}. We can find that the proposed BBDM performs robustly w.r.t. different levels of latent features. The latent space learned with downsampling factor $16$ leads to more abstract feature, as a result, the performance of the LDM model drops dramatically especially with the FID metric. 
\begin{table}[!ht]
    \centering
    \begin{tabular}{cccc}
         \bottomrule
         model & FID $\downarrow$ & LPIPS $\downarrow$ & Diversity $\uparrow$ \\
         \toprule
         LDM-f4 & 22.816 & 0.371 & 20.304 \\
         LDM-f8 & 24.530 & 0.418 & 41.625 \\
         LDM-f16 & 56.404 & 0.416 & 22.112 \\
         BBDM-f4 & 21.350 & 0.370 & 29.859 \\
         BBDM-f8 & 21.966 & 0.392 & 38.978 \\
         BBDM-f16 & 22.061 & 0.391 & 40.120 \\
         \bottomrule
    \end{tabular}
    \caption{Quantitative scores of LDM and BBDM with different downsampling factors.}
    \label{table_LDM_BBDM}
\end{table}
To further verify the image-to-image translation process during the diffusion of Brownian Bridge, we decode the latent code at each time step in the inference processes by the decoder of $VQGAN_B$. As shown in Figure~\ref{fig_latent_visual}, the input image is smoothly and gradually translated to the target domain within the Brownian Bridge.

\textbf{Sampling Steps}: 
To evaluate the influence of sampling steps in the reverse diffusion process to the performance of BBDM, we evaluate the performance with different numbers of sampling steps. In Table \ref{table_sample_steps}, we report the quantitative scores of semantic-to-image task with models trained on CelebAMask-HQ.  We can find that when the number of sampling steps is relatively small (fewer than 200 steps), the sample quality and diversity improve rapidly with the increase of sampling steps. When the number of sampling steps is relatively large (greater than 200 steps), the FID and diversity metrics get better slightly and the LPIPS metric almost remains the same as the sampling steps are raised.
\begin{table}[!ht]
    \centering
    \begin{tabular}{cccc}
         \bottomrule
         Sampling Steps & FID $\downarrow$ & LPIPS $\downarrow$ & Diversity $\uparrow$ \\
         \toprule
         20 steps & 33.409 & \textbf{0.362} & 17.587 \\
         50 steps & 25.188 & 0.372 & 23.191 \\
         100 steps & 23.503 & 0.378 & 26.157 \\
         200 steps & 21.350 & 0.370 & 29.859 \\
         1000 steps & \textbf{21.348} & 0.375 & \textbf{29.924} \\
         \bottomrule
    \end{tabular}
    \caption{Quantitative scores of different numbers of sampling steps on CelebAMask-HQ.}
    \label{table_sample_steps}
\end{table}
\textbf{The Influence of maximum variance of Brownian Bridge.}
As shown in Eq.\ref{Eq-control_variance}, we can control the diversity of Brownian Bridge through scaling the maximum variance of Brownian Bridge which can be achieved at $t = \frac{T}{2}$ by a factor $s$. In this section, we conduct several experiments taken on $s \in \{1, 2, 4\}$ to investigate the influence of $s$ to the performance of our Brownian Bridge model.
The quantitative metrics are shown in Table~\ref{table_maximum_variance}. With the increase of $s$, the diversity grows but the quality and fidelity decrease. This phenomenon 
is consistent with the observation
in Section~\ref{section 3.1} that if we use the original variance design of Brownian Bridge, BBDM cannot generate reasonable samples due to the extremely large maximum variance.
\begin{table}[!ht]
    \centering
    \begin{tabular}{cccc}
         \bottomrule
         $s$ & FID $\downarrow$ & LPIPS $\downarrow$ & Diversity $\uparrow$ \\
         \toprule
         $s=0.5$ & 22.627 & 0.387 & 27.791 \\
         $s=1$ & 21.350 & 0.370 & 29.859 \\
         $s=2$ & 23.278 & 0.380 & 37.063 \\
         $s=4$ & 24.490 & 0.384 & 39.573 \\
         \bottomrule
    \end{tabular}
    \caption{Quantitative scores of different factor $s$ on CelebAMask-HQ.}
    \label{table_maximum_variance}
\end{table}

\section{Conclusion and Future Work}
We proposed a new method for image-to-image translation based on Brownian Bridge. Compared with other diffusion-based methods, the proposed BBDM framework learns the translation between two domains directly through the Brownian Bridge diffusion process rather than a conditional generation process. We showed that our BBDM framework can generate promising results on several different tasks. Nevertheless, there is still much room for improvement of BBDM, e.g., it would be interesting to apply our framework to various multi-modal tasks like text-to-image.

\section*{Acknowledgments}
The work was funded by Natural Science Foundation of China (NSFC) under Grant 62172198, 61762064, Key Project of Jiangxi Natural Science Foundation 20224ACB202008, and the Opening Project of Nanchang Innovation Institute , Peking University.

\newpage

{\small
\bibliographystyle{ieee_fullname}
\bibliography{egbib}
}

\newpage

\appendix
\onecolumn

This supplementary material provides details that are not included in the main paper due to space limitations. We first fill in the deduction details of Section~\ref{section3.1.3}. Then the implementation details of BBDM will be provided. After that we will provide the user study results. Finally, we will present more qualitative experiment results. The code is publicly available at \href{https://github.com/xuekt98/BBDM}{https://github.com/xuekt98/BBDM}

\section{Deduction Details of Training Objective}
As shown in Eq.(~\ref{Eq-BBReverseMarkov}) 
\begin{align}
    q_{BB}(\bm{x}_{t-1} | \bm{x}_t, \bm{x}_0, \bm{y}) = \mathcal{N}(\bm{x}_{t-1}; \bm{\Tilde{\mu}}_t(\bm{x}_t, \bm{x}_0, \bm{y}) , \Tilde{\delta}_t \bm{I}) \notag
\end{align}
which can be written in the following Probability Density Function(PDF) form:
\begin{align}
    f(\bm{x}_{t-1}) = \frac{1}{\sqrt{2\pi \Tilde{\delta}_t}} e^{-\frac{(\bm{x}_{t-1} - \bm{\Tilde{\mu}}_t(\bm{x}_t, \bm{x}_0, \bm{y})^2}{2\Tilde{\delta}_t}} \notag 
\end{align}

In the same way, the right part of Eq.(~\ref{Eq-BBReverseMarkov}) can also be represented as PDF which contains three sub-parts.

From Eq.(~\ref{Eq-BBForwardMarkov}), we can have $q_{BB}(\bm{x}_t | \bm{x}_{t-1}, \bm{y})$:

\begin{align}
    f(\bm{x}_t) = \frac{1}{\sqrt{2\pi \delta_{t|t-1}}} e^{-\frac{\Big(\bm{x}_{t} - \big(\frac{1-m_t}{1-m_{t-1}}\bm{x}_{t-1} + (m_t - \frac{1-m_t}{1-m_{t-1}}m_{t-1})\bm{y}\big)\Big)^2}{2\delta_{t|t-1}}} \notag
\end{align}

The PDF of $q_{BB}(\bm{x}_{t-1} | \bm{x}_0, \bm{y})$ and $q_{BB}(\bm{x}_t | \bm{x}_0, \bm{y})$ can also be derived based on Eq.(~\ref{Eq-BB}):

\begin{align}
    f(\bm{x}_{t-1}) &= \frac{1}{\sqrt{2\pi \delta_{t-1}}} e^{-\frac{\Big(\bm{x}_{t-1} - \big(1-m_{t-1})\bm{x}_0 + m_{t-1}\bm{y}\big)\Big)^2}{2\delta_{t-1}}} \notag \\
    f(\bm{x}_{t}) &= \frac{1}{\sqrt{2\pi \delta_{t}}} e^{-\frac{\Big(\bm{x}_{t} - \big(1-m_{t})\bm{x}_0 + m_{t}\bm{y}\big)\Big)^2}{2\delta_{t}}} \notag
\end{align}

Considering that the PDF of the left part and right part of Eq.(~\ref{Eq-BBReverseMarkov}) should be equal, the following equation can be derived:

\begin{gather}
    \frac{1}{\sqrt{2\pi \Tilde{\delta}_t}} e^{-\frac{(\bm{x}_{t-1} - \bm{\Tilde{\mu}}_t(\bm{x}_t, \bm{x}_0, \bm{y})^2}{2\Tilde{\delta}_t}} = \frac{\frac{1}{\sqrt{2\pi \delta_{t|t-1}}} e^{-\frac{\Big(\bm{x}_{t} - \big(\frac{1-m_t}{1-m_{t-1}}\bm{x}_{t-1} + (m_t - \frac{1-m_t}{1-m_{t-1}}m_{t-1})\bm{y}\big)\Big)^2}{2\delta_{t|t-1}}} \frac{1}{\sqrt{2\pi \delta_{t-1}}} e^{-\frac{\Big(\bm{x}_{t-1} - \big(1-m_{t-1})\bm{x}_0 + m_{t-1}\bm{y}\big)\Big)^2}{2\delta_{t-1}}}}{\frac{1}{\sqrt{2\pi \delta_{t}}} e^{-\frac{\Big(\bm{x}_{t} - \big(1-m_{t})\bm{x}_0 + m_{t}\bm{y}\big)\Big)^2}{2\delta_{t}}}} \notag \\
    = \frac{1}{\sqrt{2\pi}}{\sqrt{\frac{\delta_{t}}{\delta_{t|t-1} \delta_{t-1}}}} e^{-\frac{1}{2\frac{\delta_{t|t-1} \delta_{t-1}}{\delta_{t}}} \Big( \bm{x}_{t-1} - \big( \frac{\delta_{t-1}}{\delta_t}\frac{1-m_t}{1-m_{t-1}}\bm{x}_t + (1-m_{t-1}\frac{\delta_{t|t-1}}{\delta_t})\bm{x}_0 + (m_{t-1} - m_t\frac{1-m_t}{1-m_{t-1}}\frac{\delta_{t-1}}{\delta_t})\bm{y} \big) \Big)^2} \notag
\end{gather}

Then we can have the following equations:
\begin{align}
    \bm{\Tilde{\mu}}_t(\bm{x}_t, \bm{x}_0, \bm{y}) = \frac{\delta_{t-1}}{\delta_t}\frac{1-m_t}{1-m_{t-1}}\bm{x}_t + (1-m_{t-1}\frac{\delta_{t|t-1}}{\delta_t})\bm{x}_0 \notag + (m_{t-1} - m_t\frac{1-m_t}{1-m_{t-1}}\frac{\delta_{t-1}}{\delta_t})\bm{y} \notag
\end{align}

\begin{align}
    \Tilde{\delta}_t = \frac{\delta_{t|t-1}\cdot\delta_{t-1}}{\delta_t} \notag
\end{align}
which is equivalent to Eq.(~\ref{Eq-BBMean}) and Eq.(~\ref{Eq-BBVar}).

\section{Implementation Details}
In this section, we provide more implementation details of BBDM, including network hyperparameters and optimization, details of training and sampling procedures.

\textbf{Network hyperparameters}. As mentioned in Section~\ref{section4.1}, we adopt the same VQGAN model and network architecture as LDM for fair comparison. In order to enable the model to be trained on single GeForce GTX 3090 GPU, we reduced model size by modifying the total number of middle layers and channels of middel features. The network details are shown in Table~\ref{table_network_parameters}.

\begin{table*}[ht]
    \setlength{\tabcolsep}{1.5mm}{
    \centering
    \begin{tabular}{ccccccc}
    \bottomrule
    model & z-shape & channel multiplier & attention resolutions & channels & total parameters & trainable parameters \\
    \toprule
    BBDM-f4 & $64 \times 64 \times 3$ & 1,4,8 & 32,16,8 & 128 & 292.42M & 237.09M \\
    BBDM-f8 & $32 \times 32 \times 4$ & 1,4,8 & 32,16,8 & 128 & 304.81M & 237.10M \\
    BBDM-f16 & $16 \times 16 \times 8$ & 1,4,8 & 16,8,4 & 128 & 327.71M & 258.11M\\
    \bottomrule
    \end{tabular}
    }
    \caption{Network hyperparameters for both BBDM and LDM used in this paper.}
    \label{table_network_parameters}
\end{table*}

\textbf{Training and sampling details}. In order to improve the performance of BBDM, Exponential Moving Average(EMA) was adopted in the training procudure together with ReduceLROnPlateau learning rate scheduler.

\begin{table*}[ht]
    \centering
    \begin{tabular}{ccccc}
    \bottomrule
    model & EMA start step & EMA decay & EMA update interval & batch size \\
    \toprule
    BBDM-f4 & 30000 & 0.995 & 16 & 8  \\
    BBDM-f8 & 30000 & 0.995 & 16 & 8  \\
    BBDM-f16 & 15000 & 0.995 & 8 & 16 \\
    \bottomrule
    \end{tabular}
    \captionof{table}{EMA hyperparameters of BBDM.}
    \label{table_EMA_parameters}
\end{table*}

\begin{table*}[ht]
    \centering
    \begin{tabular}{ccccccc}
    \bottomrule
    model & max learning rate & min learning rate & factor & patience & cool down & threshold \\
    \toprule
    BBDM-f4 & 1.0e-4 & 5.0e-7 & 0.5 & 3000 & 2000 & 1.0e-4  \\
    BBDM-f8 & 1.0e-4 & 5.0e-7 & 0.5 & 3000 & 2000 & 1.0e-4  \\
    BBDM-f16 & 1.0e-4 & 1.0e-6 & 0.5 & 3000 & 2000 & 1.0e-4 \\
    \bottomrule
    \end{tabular}
    \captionof{table}{Learning rate scheduler hyperparameters of BBDM.}
    \label{table_LRScheduler_parameters}
\end{table*}

\begin{figure}[!ht]
\begin{center}
    \includegraphics[width=0.5\linewidth]{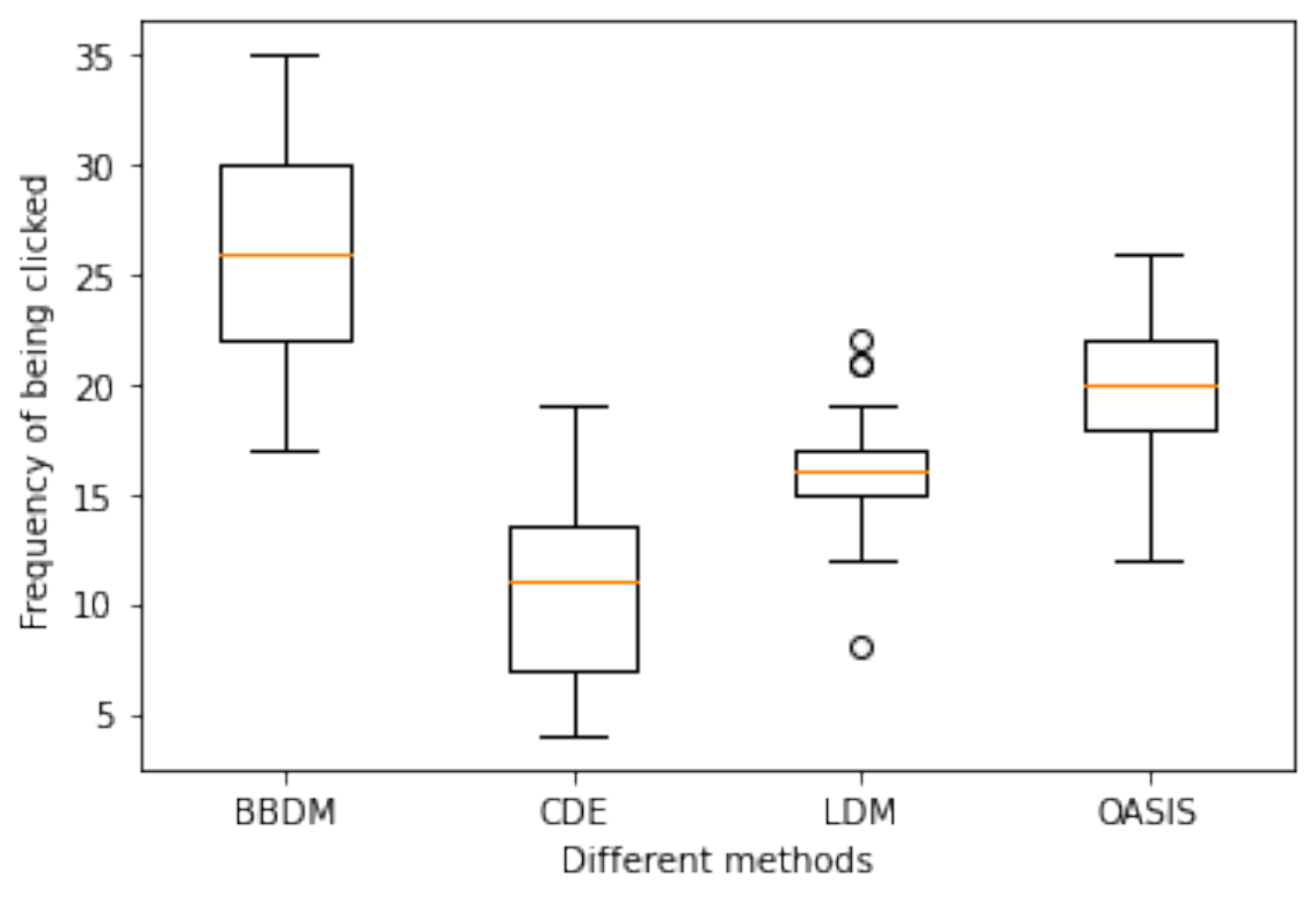}
\end{center}
  \caption{User study results of BBDM, CDE, LDM, OASIS on CelebAMask-HQ dataset.}
\label{fig_user_study_appendix}
\end{figure}

\section{User Study}
An additional subjective user study is designed to evaluate the performance of the proposed method against three methods with comparable FID measurement, including CDE, LDM and OASIS. 
12 groups of samples are randomly selected from CelebAMask-HQ experiments. 
For each sample, a pair of editing results are randomly shown to the participants. As there are 4 different editing results for each image, 72 clicks are required for each participant. 112 users with age between 20 and 50 were invited to participate in the user study. The distribution of user preference is shown in Figure~\ref{fig_user_study_appendix}. We can see that more users prefer the results of the proposed method.


\section{Additional Qualitative Results}
\begin{figure}[!ht]
\begin{center}
    \includegraphics[width=\linewidth]{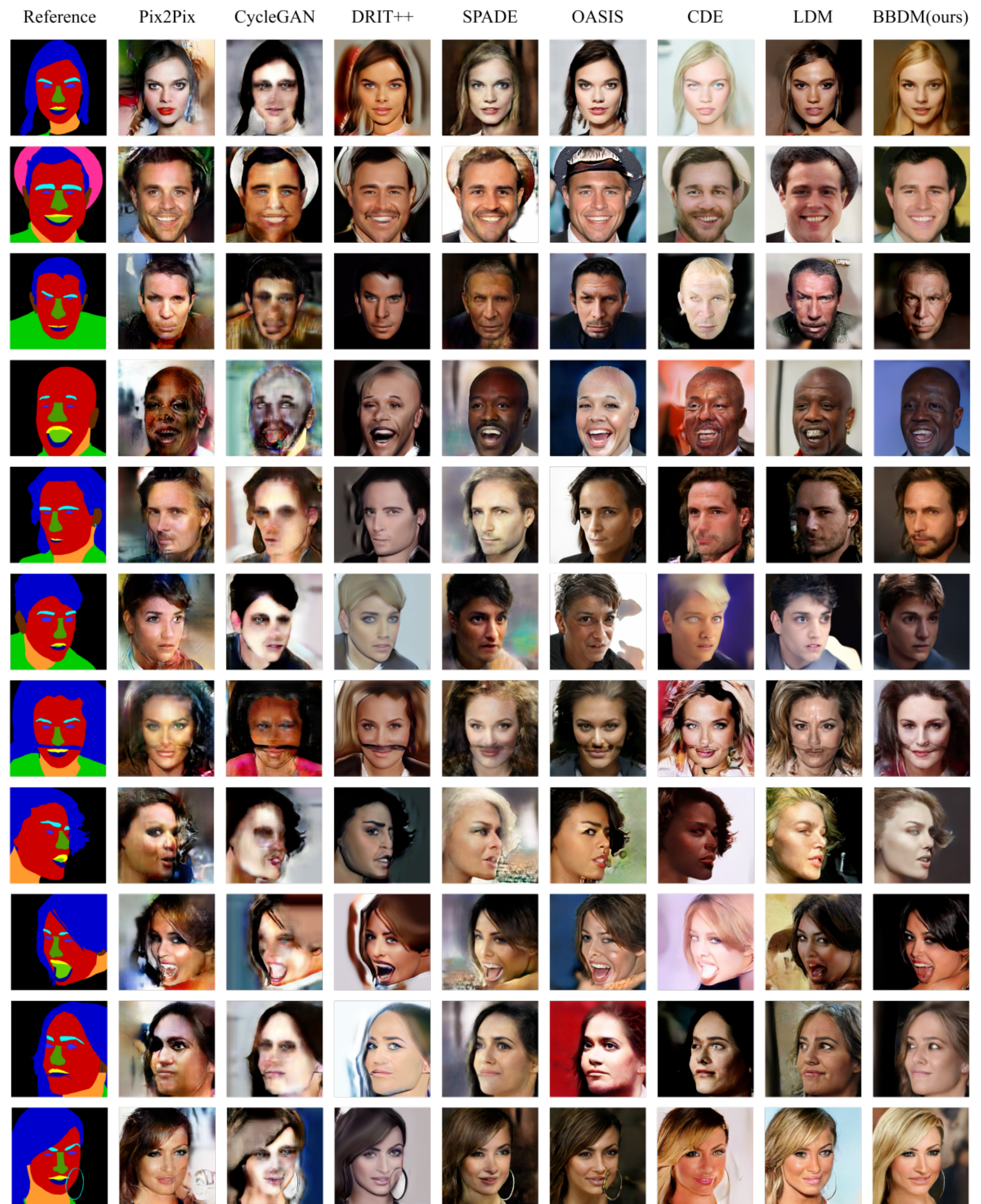}
\end{center}
  \caption{More qualitative results on the CelebAMask-HQ dataset.}
\label{fig_BBDM_CelebAMaskHQ_appendix}
\end{figure}

\begin{figure}[!ht]
\begin{center}
    \includegraphics[width=0.9\linewidth]{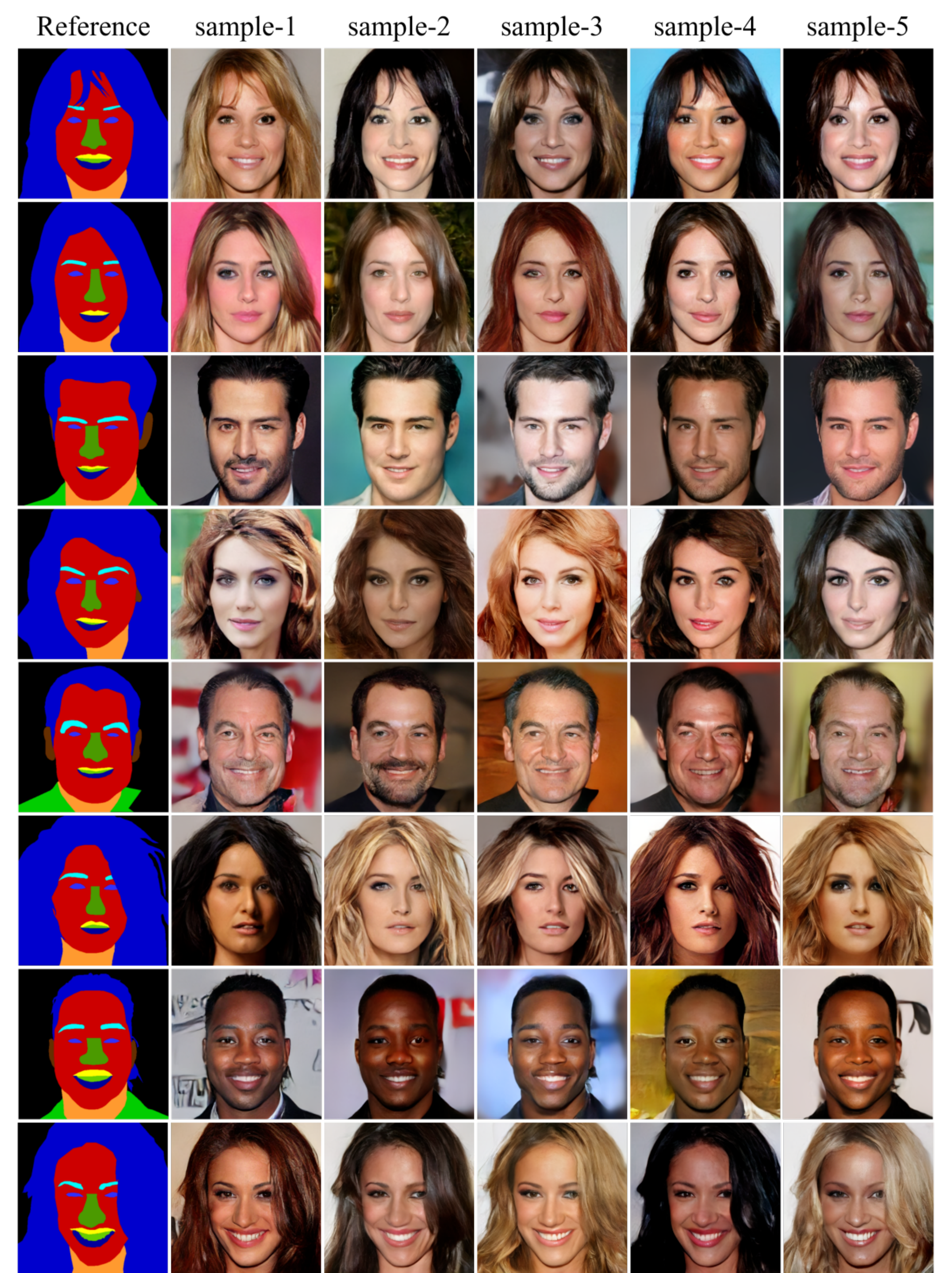}
\end{center}
  \caption{More diverse samples on the CelebAMask-HQ dataset.}
\label{fig_BBDM_CelebAMaskHQ_diversity_appendix}
\end{figure}

\begin{figure}[!ht]
\begin{center}
    \includegraphics[width=0.9\linewidth]{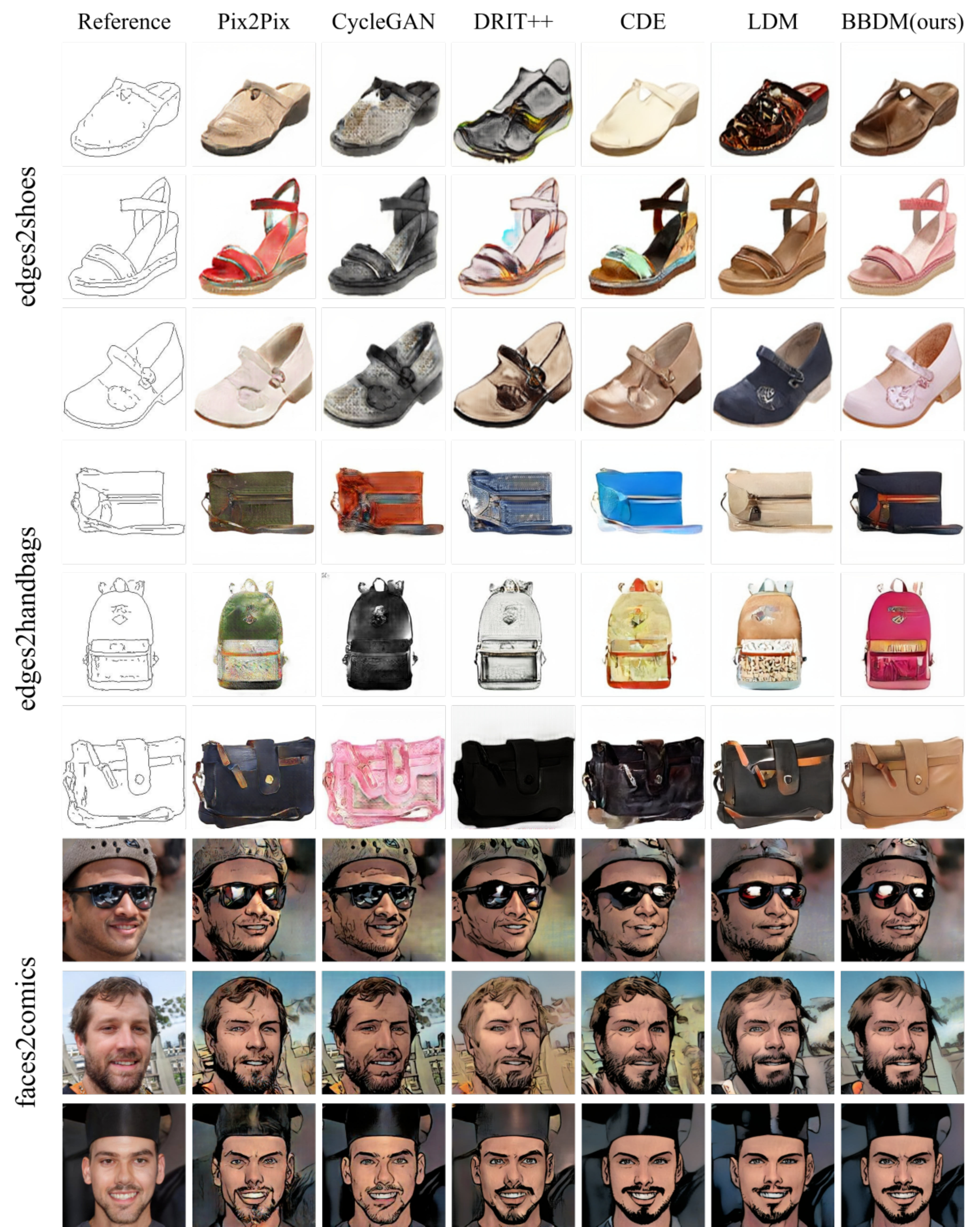}
\end{center}
  \caption{More qualitative results on the edges2shoes, edges2handbags and faces2comics datasets.}
\label{fig_BBDM_quality_appendix}
\end{figure}

\begin{figure}[!ht]
\begin{center}
    \includegraphics[width=0.9\linewidth]{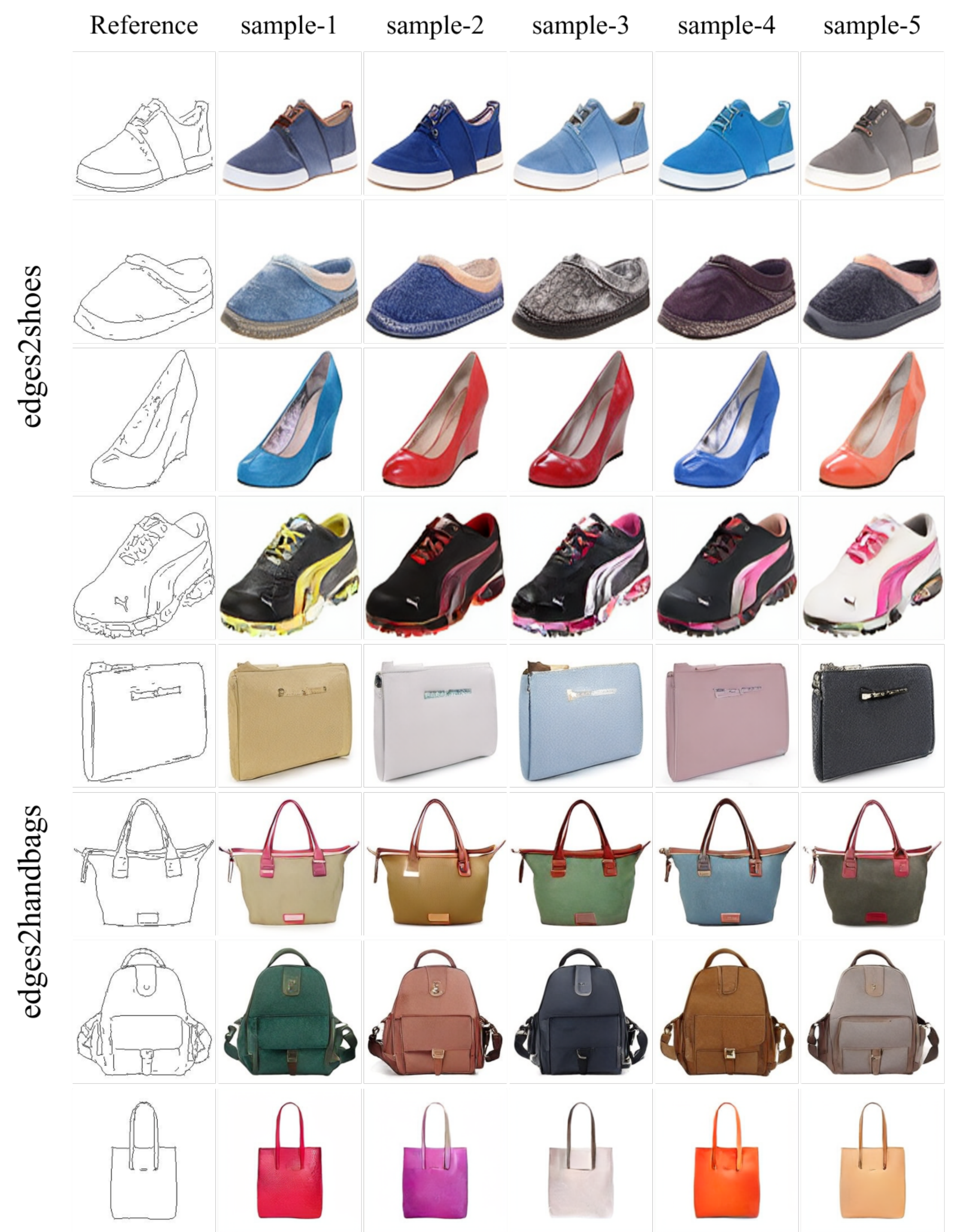}
\end{center}
  \caption{More diverse samples on the edges2shoes, edges2handbags datasets.}
\label{fig_BBDM_diversity_appendix}
\end{figure}

\begin{figure}[!ht]
\begin{center}
    \includegraphics[width=0.8\linewidth]{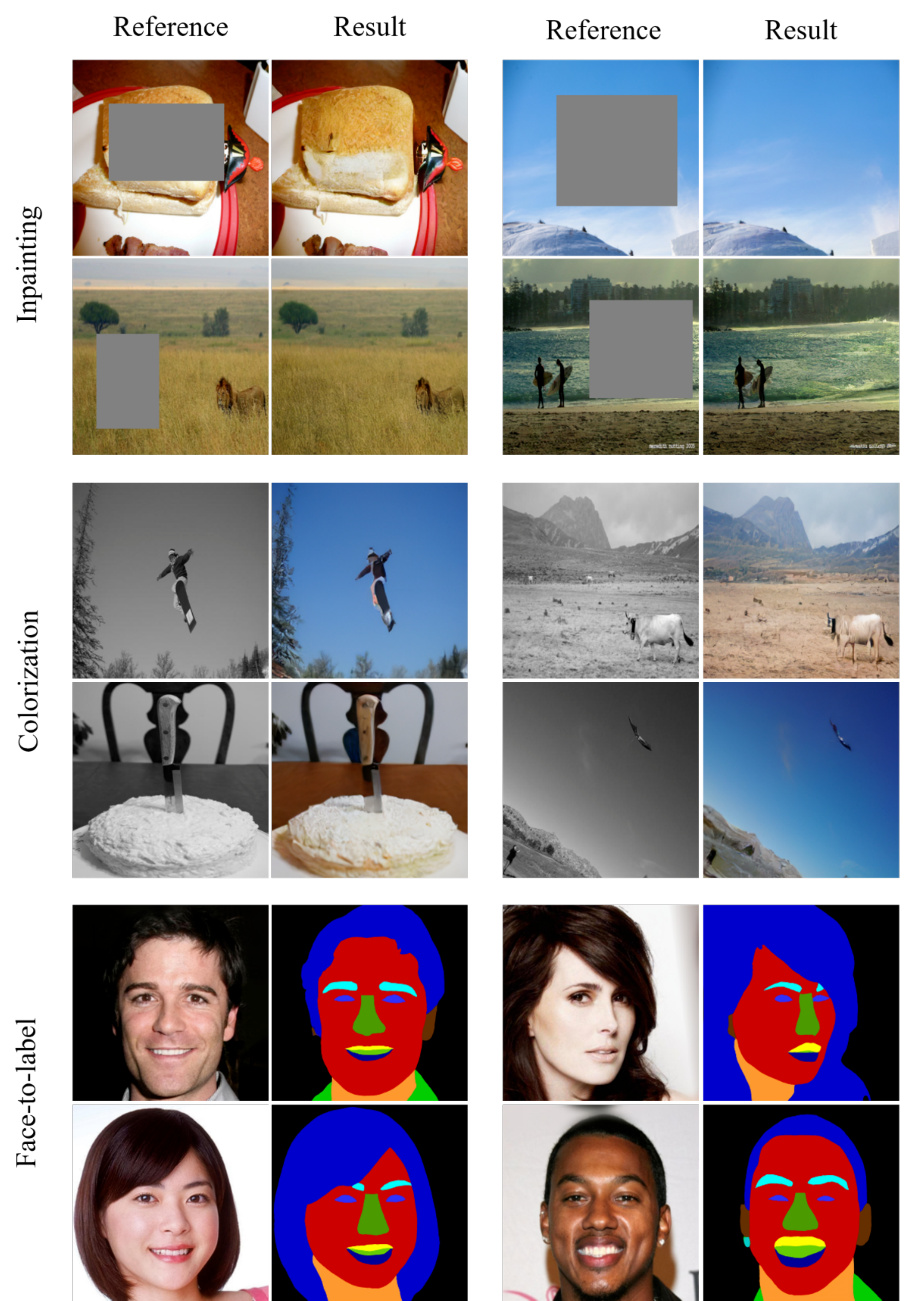}
\end{center}
  \caption{More inpainting, colorization and face-to-label samples.}
\label{fig_BBDM_other_appendix}
\end{figure}

Finally, we provide additional qualitative results compared with other challending methods (Figure~\ref{fig_BBDM_CelebAMaskHQ_appendix},~\ref{fig_BBDM_quality_appendix}). More diverse samples are shown in Figure~\ref{fig_BBDM_CelebAMaskHQ_diversity_appendix} and ~\ref{fig_BBDM_diversity_appendix}. Other experiment results on inpainting, colorization and face-to-label tasks can be found in Figure~\ref{fig_BBDM_other_appendix}.

\newpage
\end{document}